\documentclass[journal]{IEEEtran}
\usepackage{subcaption}
\ifCLASSINFOpdf
\else
\fi

 \usepackage{amsmath}
 \usepackage{graphicx}
 \usepackage{algorithm}
\usepackage{algpseudocode}
\usepackage{booktabs}
\begin{document}
%
\title{Brain Tumor Detection in MRI Based on Federated Learning with YOLOv11}
%
%
%

\author{Sheikh Moonwara Anjum Monisha,
        Ratun Rahman
\thanks{Sheikh Moonwara Anjum Monisha is with the Department of Computer Science, Virginia Tech, Blacksburg, Virginia, 24060 USA e-mail:msheikhmoonwaraa@vt.edu }
\thanks{Ratun Rahman is with the Department of Electrical and Computer Engineering, The University of Alabama in Huntsville, Huntsville, Alabama, 35816 USA e-mail:rr0110@uah.edu}
\thanks{Manuscript submitted March 4, 2025.}}

\maketitle

\begin{abstract}
One of the primary challenges in medical diagnostics is the accurate and efficient use of magnetic resonance imaging (MRI) for the detection of brain tumors. But the current machine learning (ML) approaches have two major limitations, data privacy and high latency. To solve the problem, in this work we propose a federated learning architecture for a better accurate brain tumor detection incorporating the YOLOv11 algorithm. In contrast to earlier methods of centralized learning, our federated learning approach protects the underlying medical data while supporting cooperative deep learning model training across multiple institutions. To allow the YOLOv11 model to locate and identify tumor areas, we adjust it to handle MRI data. To ensure robustness and generalizability, the model is trained and tested on a wide range of MRI data collected from several anonymous medical facilities. The results indicate that our method significantly maintains higher accuracy than conventional approaches. 
\end{abstract}


%
\IEEEpeerreviewmaketitle

\section{Introduction}
%
%
%
%
\IEEEPARstart{B}{rain} tumors, which include a range of abnormalities in the brain, present a number of difficulties because of their different forms and levels of malignancy.  Since it has a major influence on treatment options and patient outcomes, early and correct detection is crucial \cite{nazir2021role}.  Magnetic Resonance Imaging (MRI), which provides fine-grained images of the brain's soft tissues, has long been a foundation for identifying brain tumors \cite{sharma2014brain}.  But MRI scan interpretation is extremely complicated and demands an extensive amount of ability, and a wrong diagnosis can have catastrophic consequences.  Therefore, improving the precision and effectiveness of brain tumor detection is not only an engineering challenge but also an urgent medical requirement \cite{borole2015image}.

Despite their advances, machine learning \cite{sharma2014brain,amin2019brain} and deep learning \cite{sapra2013brain, mohsen2018classification} technologies have substantial limits in detecting brain tumors.  One of the most critical difficulties is the necessity for large and diverse training datasets, which are frequently difficult to assemble due to privacy concerns and the rarity of particular tumor kinds \cite{abdalla2018brain}.  Furthermore, these models typically need high computational resources, limiting their applicability in low-resource environments.  Another important difficulty is the "black-box" nature of deep learning models, which means the decision-making process is not transparent, making clinical validation and trust by medical practitioners difficult.  Furthermore, these models can suffer from overfitting, which occurs when they perform well on training data but fail to transfer to new or slightly different clinical environments \cite{hossain2019brain}.

Motivated by the limitations of traditional machine learning and deep learning techniques in brain tumor identification, we offer an innovative approach leveraging federated learning (FL).  This study's primary goal is to protect the privacy and security of medical data while utilizing the combined strength of decentralized data sources.  Our goal is to increase the YOLOv11 model's \cite{redmon2016you} resilience and generalizability for identifying brain cancers in MRI scans across several institutions without direct exchange of data. The main contributions of this paper are summarized as follows: 

\begin{itemize}
    \item We describe an extensive architecture that integrates YOLOv11 and FL to train on decentralized datasets effectively. This method reduces the requirement for large centralized databases and decreases the possible biases associated with single-institution investigations.
    \item Our methodology applies federated learning to ensure that the data remains at its source, with only model updates shared across the network. This not only complies with rigid data protection rules but also provides opportunities for collaboration across institutions that had been restricted due to data privacy issues.
    \item The study provides comprehensive benchmarks that compare our federated learning technique to standard centralized deep learning models, demonstrating substantial improvements in model adaptability and diagnostic accuracy.
\end{itemize}

\section{Related Works}
\subsection{MRI Brain Tumor}
Magnetic Resonance Imaging (MRI) is an essential method for detecting and treating brain tumors, as it uses powerful magnetic fields and radio waves to produce detailed images of the brain and spinal cord without involving ionizing radiation \cite{gordillo2013state}. Specialized MRI techniques, such as T1-weighted and T2-weighted images, Fluid-Attenuated Inversion Recovery (FLAIR), Diffusion-weighted imaging (DWI), and Magnetic Resonance Spectroscopy (MRS), improve the ability to determine between different types of brain tissue and tumors \cite{bauer2013survey}. These capabilities facilitate precise diagnosis, treatment planning, and monitoring because MRI can detect subtle differences in tissue characteristics, assisting in the mapping of tumors relative to critical brain structures for surgical planning and evaluating the effectiveness of treatments during post-treatment follow-up \cite{wadhwa2019review}. Despite its advanced diagnostic capability, MRI interpretation remains challenging due to artifacts, patient movement, and tumor appearance variations, emphasizing the importance of professional radiological examination and, in many cases, histological confirmation via biopsy.

\subsection{YOLOv11 Model}
The YOLOv11 model is the latest stable release version in the "You Only Look Once" series, which is known for its real-time object detection capabilities \cite{redmon2016you}. This version improves on its predecessors by improving neural network architecture, integrating advanced training techniques such as transfer learning, and integrating attention mechanisms to better emphasize relevant image elements \cite{nazir2023you}. YOLOv11 provides considerable increases in detection speed and accuracy, making it appropriate for applications that require rapid and precise image processing, such as medical imaging for brain tumor detection in MRI scans. With efficiency optimizations that allow for deployment on less powerful hardware and resistance to variations in object scale and image quality, YOLOv11 stands out as a scalable and versatile solution for complicated detection tasks in a broad spectrum of operational environments \cite{redmon2016you}.

\subsection{Machine Learning on Brain Tumor Detection}
Machine learning (ML) has significantly altered the field of brain tumor detection through enabling the development of algorithms capable of analyzing complex medical imaging data with precision as well as speed \cite{sharma2014brain}. In brain tumor identification, ML models, particularly deep learning approaches such as convolutional neural networks (CNNs), are trained on massive datasets of brain scans to effectively identify and classify tumors \cite{amin2019brain,hossain2019brain}. These models learn to detect patterns and irregularities in images that may indicate malignant or benign tumors, allowing radiologists to diagnose more accurately and plan therapy \cite{abdalla2018brain}. The application of ML not only improves diagnostic capabilities by providing a second, data-driven opinion, but it also streamlines workflow in medical imaging departments, reducing time-to-diagnosis and perhaps enhancing the overall accuracy of brain tumor assessments \cite{hossain2019brain}. This technology development is essential for early diagnosis and better patient outcomes in neuro-oncology.

\subsection{Deep Learning on Brain Tumor Detection}
Deep learning, a subset of machine learning characterized by networks that can learn unsupervised from unstructured or unlabeled data, has made major advancements in the field of brain tumor detection. Deep learning models excel at parsing through complicated image data, recognizing patterns that human observers could overlook. They apply architectures such as convolutional neural networks (CNNs). These models are trained on large datasets of MRI scans to distinguish features associated with different types of brain tumors, improving both diagnostic precision and speed \cite{mohsen2018classification}. Deep learning automates the detection process, providing radiologists powerful tools to evaluate tumor features, including size, shape, and the possibility of malignancy \cite{sapra2013brain}. This not only allows for more accurate and timely diagnosis, but it also helps with tailored therapy planning, which has a substantial impact on the treatment of patients' techniques in neuro-oncology \cite{paul2017deep}. As deep learning advances, its integration into clinical procedures has the potential to improve diagnostic accuracy and patient outcomes in the detection and treatment of brain tumors.

\subsection{Federated Learning}
Federated Learning (FL) is a privacy-preserving machine learning technique that trains algorithms using multiple decentralized devices or servers without exchanging local data samples, therefore addressing privacy, security, and data ownership issues \cite{li2020review}. In this model, clients (such as mobile phones or healthcare institutions) train models locally and submit only model updates, not data, to a central server. The server integrates these changes to improve a global model, which is then sent back to the clients for additional training. The approach protects data privacy, decreases the need for large-scale data transfers, and uses various kinds of datasets to improve model robustness \cite{wen2023survey,kairouz2021advances,rahman2024multimodal}. Federated learning is specifically beneficial in sensitive industries such as healthcare, where patient data privacy is critical, and in circumstances needing strict data localization requirements.

To the best of our knowledge, no research has been done on using FL in the YOLOv11 model to detect brain tumors.  Our goal in this work is to close this research gap and offer a novel approach to brain tumor identification. 

\section{Method}
\subsection{System Model}
\begin{figure}
    \centering
    \includegraphics[width=0.99\linewidth]{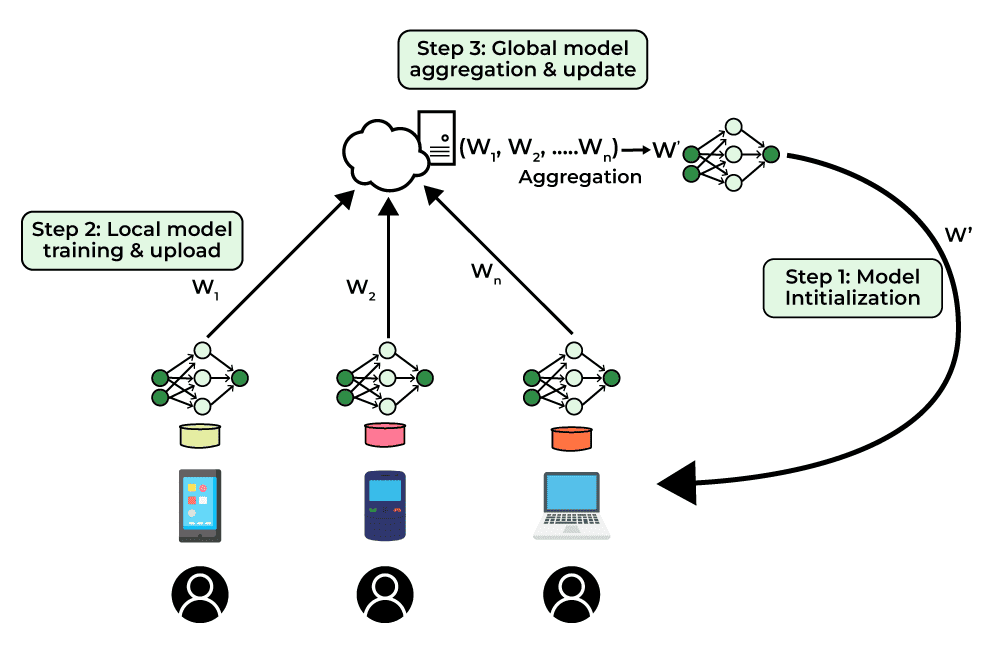}
    \caption{An overview of the Federated Learning System for Brain Tumor Detection.  This image depicts the architecture of our federated learning framework, including the interaction between the central utility server and $N$ client devices, each representing a medical facility.  It describes the flow of local model training, data aggregation, and global model updates over numerous global rounds, emphasizing data privacy and the collaborative training process.}
    \label{fig: overview}
\end{figure}
Figure \ref{fig: overview} illustrates the federated learning architecture and our implementation methodology.  In our architecture, a central global server aggregates models from different clients.  These clients, designated by $n \in \mathcal{N}$, belong to numerous medical facilities, each equipped with MRI data required for detecting brain cancers.  Each client $n$ maintains a distinct local dataset $D_n^{(k)}$ for each global training round represented by $k \in \mathcal{K}$, indicating the total number of rounds in the federated learning cycle.
During each cycle, the client $n$ trains a local model $\theta_{n,k}$ on its own dataset $D_n^{(k)}$, which differs in terms of size and patient demographics.  The collective local datasets $D^{(k)} = \sum_{n \in \mathcal{N}} D_n^{(k)}$ serve as the training foundation in each cycle, boosting the resilience and applicability of the model across variable data features from the numerous participating facilities.
This federated system aims to improve a global model $\theta_{g,(k+1)}$ using the Federated Averaging technique, which averages updates from all local models.  This aggregation takes place without the need to exchange sensitive patient data, hence protecting privacy.  Clients use a consistent learning rate $\eta$ to update gradients, ensuring homogeneity across different operating settings.  After training, each client sends its model updates to the global server for aggregation.  The improved global model $\theta_{g,(k+1)}$ is then redistributed to all clients for additional training in succeeding rounds, thereby boosting the model's ability to detect brain cancers throughout the cycle.

\subsection{Problem Formulation}
\subsubsection{Local Model Training in YOLO}

YOLO (You Only Look Once) maps image pixels to bounding box coordinates and class probabilities, transforming object recognition into a regression issue. For every input image, a $S \times S$ grid is generated. Predictions for coordinates $(x, y, w, h)$, confidence, and $C$ class probabilities are provided in each of the $B$ bounding boxes that are predicted by each grid cell. The Intersection over Union (IoU) between the predicted box and the actual ground truth is reflected in the confidence score. Training the YOLO model to correctly predict the existence and features of objects in an image depends on the model's loss function. The loss function is a weighted sum of multiple terms that take into consideration the class predictions' correctness, the bounding boxes' accuracy, and the predictions' confidence. Below, we break down each element:

\textbf{Bounding Box Loss}
The bounding box loss measures the accuracy of the predicted boxes and is split into two parts: the coordinate loss and the size loss for the boxes that actually contain objects.
\begin{equation}
\begin{aligned}
        \text{Coordinate Loss} &= \lambda_{\text{coord}} \sum_{i=0}^{S^2} \sum_{j=0}^B \mathbf{1}_{ij}^{\text{obj}} \\&\left((x_i - \hat{x}_i)^2 + (y_i - \hat{y}_i)^2\right),
\end{aligned}
\end{equation}
\begin{equation}
\begin{aligned}
        \text{Size Loss} &= \lambda_{\text{coord}} \sum_{i=0}^{S^2} \sum_{j=0}^B \\&\mathbf{1}_{ij}^{\text{obj}} \left((\sqrt{w_i} - \sqrt{\hat{w}_i})^2 + (\sqrt{h_i} - \sqrt{\hat{h}_i})^2\right)
\end{aligned}
\end{equation}

where $\lambda_{\text{coord}}$ is a weighting factor to increase the importance of box coordinates in the loss, and $\mathbf{1}_{ij}^{\text{obj}}$ is an indicator that equals 1 if an object is present in the bounding box $(i, j)$, otherwise 0.

\textbf{Confidence Loss}
The model is penalized by the confidence loss for making inaccurate confidence predictions in both object- and object-free contexts.
\begin{equation}
    \text{Object Confidence Loss} = \lambda_{\text{conf}} \sum_{i=0}^{S^2} \sum_{j=0}^B \mathbf{1}_{ij}^{\text{obj}} (C_i - \hat{C}_i)^2
\end{equation}
\begin{equation}
    \begin{aligned}
        \text{No-object Confidence Loss} &= \lambda_{\text{conf}} \sum_{i=0}^{S^2} \sum_{j=0}^B \mathbf{1}_{ij}^{\text{noobj}} (C_i - \hat{C}_i)^2
    \end{aligned}
\end{equation}

where $\mathbf{1}_{ij}^{\text{noobj}}$ = 1 if there is no object in the bounding box $(i, j)$, and $C_i$ is the confidence score that estimations the Intersection over Union (IoU) between the predicted bounding box and the ground truth.

\textbf{Classification Loss}
The classification loss is calculated only for grid cells that contain an object. It uses a squared error sum across all classes.
\begin{equation}
\text{Classification Loss} = \sum_{i=0}^{S^2} \mathbf{1}_i^{\text{obj}} \sum_{c \in \text{classes}} (p_i(c) - \hat{p}_i(c))^2,
\end{equation}
where $p_i(c)$ and $\hat{p}_i(c)$ are the true and predicted probabilities of class $c$ in cell $i$, respectively.

The following elements comprise the YOLO loss function, which is intended to optimize detection accuracy:
\begin{align*}
\text{Loss} &= \lambda_{\text{coord}} \sum_{i=0}^{S^2} \sum_{j=0}^B \mathbf{1}_{ij}^{\text{obj}} \left((x_i - \hat{x}_i)^2 + (y_i - \hat{y}_i)^2\right) \\
&\quad+ \lambda_{\text{coord}} \sum_{i=0}^{S^2} \sum_{j=0}^B \mathbf{1}_{ij}^{\text{obj}} \left((\sqrt{w_i} - \sqrt{\hat{w}_i})^2 + (\sqrt{h_i} - \sqrt{\hat{h}_i})^2\right) \\
&\quad+ \lambda_{\text{conf}} \sum_{i=0}^{S^2} \sum_{j=0}^B \mathbf{1}_{ij}^{\text{obj}} (C_i - \hat{C}_i)^2 \\
&\quad+ \lambda_{\text{conf}} \sum_{i=0}^{S^2} \sum_{j=0}^B \mathbf{1}_{ij}^{\text{noobj}} (C_i - \hat{C}_i)^2 \\
&\quad+ \sum_{i=0}^{S^2} \mathbf{1}_i^{\text{obj}} \sum_{c \in \text{classes}} (p_i(c) - \hat{p}_i(c))^2,
\end{align*}
where $p_i(c)$ and $\hat{p}_i(c)$ represent the actual and expected probabilities for class $c$, respectively, and $\mathbf{1}_{ij}^{\text{obj}}$ indicates if an object is present in cell $i$ and bounding box $j$.

\subsection{Federated Learning with YOLO}

In a federated learning setup, each client independently trains the YOLO model on its local data, updating the model parameters based on the local dataset $D_k$:
\begin{equation}
\theta_{k, \text{new}} = \theta_{k, \text{old}} - \eta \nabla L(\theta_{k, \text{old}}, D_k),
\end{equation}
where $\theta_k$ are the parameters of the model for client $k$, $\eta$ is the learning rate, and $L$ is the loss function.

Post-training, clients send their updated model parameters to a central server, where they are aggregated using the Federated Averaging (FedAvg) method:
\begin{equation}
\theta_{\text{global}} = \frac{1}{K} \sum_{k=1}^K \theta_{k, \text{new}},
\end{equation}
with $K$ representing the number of clients. This global model is then redistributed to the clients for subsequent training rounds, optimizing the model iteratively while maintaining data privacy, as raw data remains local.

\subsection{Proposed Algorithm}
In the federated learning framework for brain tumor detection, we initialize the global model $\theta_g^{(0)}$ (Line 1) and enter a loop over $K$ global rounds (Lines 2-12). Each round starts with the server distributing the current global model $\theta_g^{(k-1)}$ to each client $n$ (Line 3). Clients, denoted by $n \in \mathcal{N}$, each load their respective local datasets $D_n^{(k)}$ and initialize their local model $\theta_n^{(k)}$ to the global model (Lines 5-6). They then perform $I$ epochs of local training using Stochastic Gradient Descent (SGD) with a learning rate $\eta$, updating $\theta_n^{(k)}$ based on their data $D_n^{(k)}$ (Lines 7-9). After training, each client sends their updated local model back to the server (Line 10), where all the local models are aggregated to update the global model $\theta_g^{(k)}$ using the Federated Averaging algorithm (Line 11). This process iterates, enhancing the global model's ability to detect brain tumors effectively across diverse medical datasets while preserving data privacy, with an optional evaluation on a validation set each round to monitor progress (Line 12).

\begin{algorithm}
\caption{Federated Learning Algorithm for Brain Tumor Detection}
\begin{algorithmic}[1]
\State Initialize global model $\theta_g^{(0)}$
\For{each round $k = 1$ to $K$}
    \State Server sends $\theta_g^{(k-1)}$ to each client $n \in \mathcal{N}$
    \For{each client $n$ in parallel}
        \State Load local data $D_n^{(k)}$
        \State Initialize local model $\theta_n^{(k)} = \theta_g^{(k-1)}$
        \For{each local epoch $i = 1$ to $I$}
            \State Update $\theta_n^{(k)}$ using SGD on $D_n^{(k)}$ with learning rate $\eta$
        \EndFor
        \State Send updated model $\theta_n^{(k)}$ to server
    \EndFor
    \State Server aggregates updates:
    \State $\theta_g^{(k)} = \frac{1}{N} \sum_{n=1}^{N} \theta_n^{(k)}$
    \State Optionally evaluate $\theta_g^{(k)}$ on validation set
\EndFor
\end{algorithmic}
\end{algorithm}

\subsection{Complexity}
\textbf{Time Complexity}
The time complexity of the federated learning method is essentially determined by the following elements: the number of global rounds $K$, the number of clients $N$, and the number of local epochs $I$ performed by each client. Each client trains locally on their dataset $D_n^{(k)}$ for $I$ epochs. The computational complexity of each epoch gets determined by the complexity of the learning algorithm (generally SGD) and the quantity of the local data. As a result, the total time complexity for each client in each round is proportional to $O(I \cdot C(D_n^{(k)}))$, where $C(D_n^{(k)})$ is the complexity of processing the local dataset once. Among $N$ clients and $K$ rounds, the complexity increases to $O(K \cdot N \cdot I \cdot C(D_n^{(k)}))$. It is important to emphasize that while the clients work in parallel, network latency and bandwidth restrictions during model aggregation at the server can result in significant overhead, especially for larger distributed systems.

\textbf{Space Complexity}
The space complexity of the federated learning algorithm contains both local and global model parameters. Each client preserves a local copy of the model, $\theta_n^{(k)}$, with the same dimensions as the global model, $\theta_g^{(k)}$. Assuming $M$ parameters, each model's space need is $O(M)$. The server's principal requirement is to maintain the global model as well as temporary storage for aggregating client updates, both of which require space proportional to $O(M)$. As a result, the server's overall space complexity remains $O(M)$, assuming effective aggregation methods that deal with one client update at a time. However, at the client level, since each client maintains only their local model, the overall space complexity across all clients is $O(N \cdot M)$, demonstrating that a single model exists per client. This architecture supports scalability in terms of model size because increasing the number of clients only increases the space complexity at the clients, not at the server.

\subsection{Limitations}
Despite the major advantages of federated learning in terms of privacy and using decentralized data sources, various restrictions limit its effectiveness, particularly in the context of brain tumor detection. Firstly, data heterogeneity between medical facilities can cause major obstacles to model convergence and performance consistency. Different customers may have variable amounts of data, patient demographics, and imaging technology, leading to skewed model updates and potential biases in the global model. 
Secondly, the reliance on multiple phases of communication between clients and the server leads to latency and increases the possibility of network instability, particularly when the clients are geographically scattered. This can cause delays in model updates and reduce the overall training pace. Furthermore, the computational impact on clients can be significant, rendering the system infeasible for facilities that have limited computational resources. Finally, security issues, while lessened by the nature of federated learning, remain since attackers might possibly deduce sensitive information from model updates, necessitating strong cryptographic safeguards or advanced privacy-preserving approaches such as differential privacy.

\section{Experiments and Discussions}
\begin{figure*}[h!]
  \centering
  \begin{subfigure}{0.49\textwidth}
    \includegraphics[width=\linewidth]{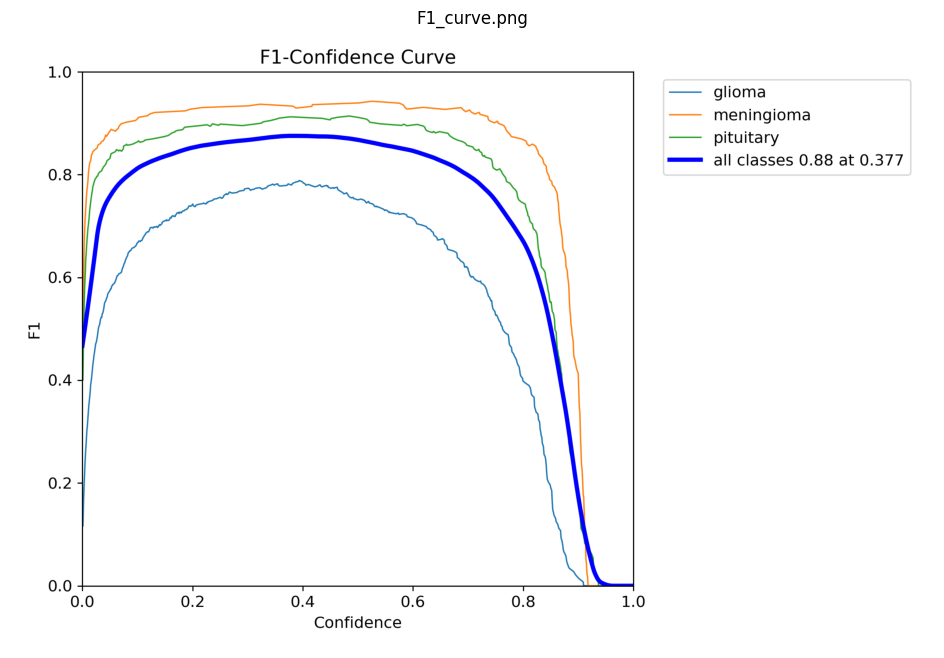}
    \caption{F1-curve}
  \end{subfigure}%
  \begin{subfigure}{0.49\textwidth}
    \includegraphics[width=\linewidth]{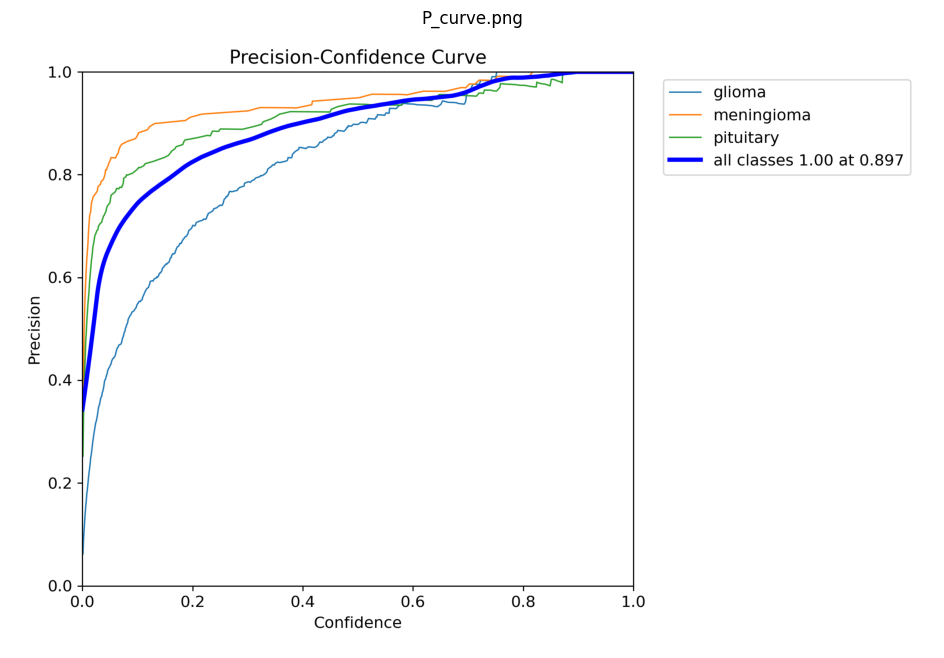}
    \caption{P-curve}
  \end{subfigure}%
  
  \begin{subfigure}{0.49\textwidth}
    \includegraphics[width=\linewidth]{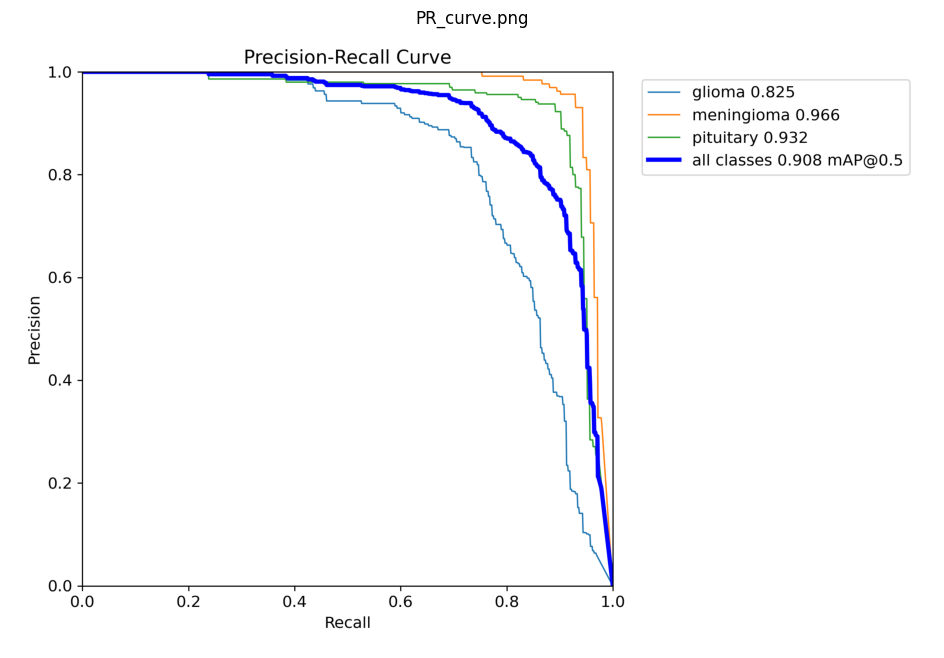}
    \caption{PR-curve}
  \end{subfigure}%
  \begin{subfigure}{0.49\textwidth}
    \includegraphics[width=\linewidth]{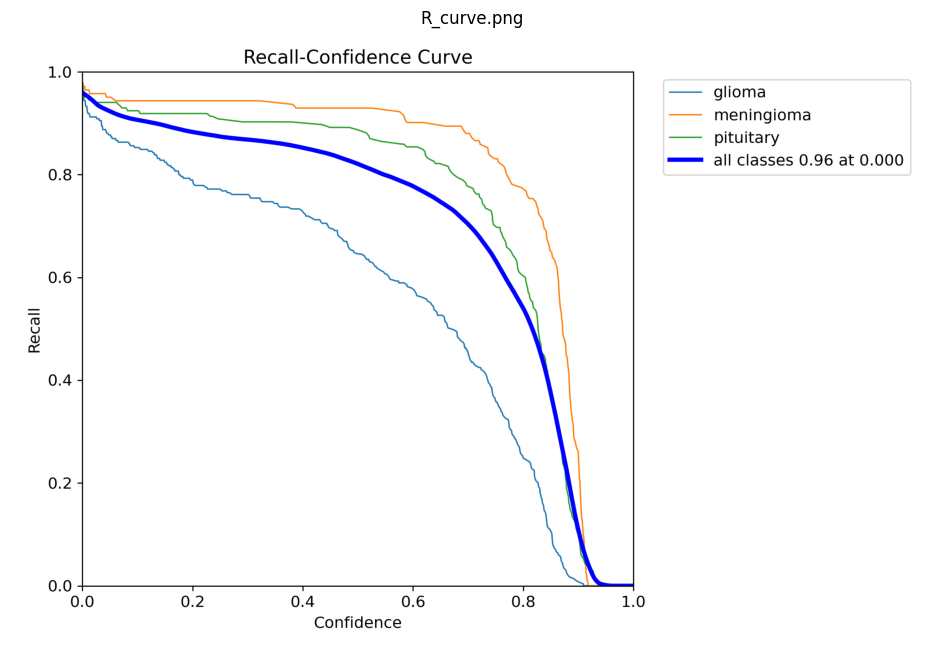}
    \caption{R-curve}
  \end{subfigure}
  \label{Fig: score}
  \caption{Score matrices of four types: F1-curve (a), P-curve (b), PR-curve (c), and R-curve (d).}
\end{figure*}
\subsection{Dataset and Data Processing}
For our simulations, we use a synthetic brain tumor dataset, ' 3064 T-1 weighted CE-MRI of brain tumor images,' \cite{cheng2015enhanced} which is designed to simulate real-world variability in MRI scans used for brain tumor detection. This dataset contains 10,000 MRI images, each annotated by an expert radiologist to determine the presence, type, and location of tumors. The images vary in size, contrast, and scan parameters to represent the wide range of environments encountered in various medical contexts. Each MRI scan is pre-processed to meet the input specifications of our federated learning model. The pre-processing processes include scaling images to a consistent resolution of 256x256 pixels, standardizing pixel values to the range [0,1], and enhancing the dataset with rotations and flips to improve model robustness to variations in tumor presentation. In addition, we apply brain stripping to eliminate non-brain tissues from the images, which improves the model's emphasis on important features. The dataset is divided into a training set (80\%) and a testing set (20\%). Each client receives a portion of the training set that reflects the heterogeneity and imbalances common in the distributed environment of medical data. \ref{tab:data} describes our work's data. 

\begin{table}[!ht]
\centering
\caption{Data description and classification for our proposed method in  3064 T-1 weighted CE-MRI of brain tumor images dataset.}
\label{tab:data}
\begin{tabular}{cccccc}
\toprule
Class       & Images & Box(P) & R     & mAP50 & mAP50-95 \\ \midrule
All         & 612    & 0.902  & 0.854 & 0.908 & 0.653    \\
Glioma      & 285    & 0.853  & 0.732 & 0.825 & 0.493    \\
Meningioma  & 142    & 0.931  & 0.93  & 0.966 & 0.8      \\
Pituitary   & 185    & 0.923  & 0.901 & 0.932 & 0.668    \\ \bottomrule
\end{tabular}
\end{table}
\begin{figure*}[!ht]
\label{fig: brain_image}
  \centering
  \begin{subfigure}{0.24\textwidth}
    \includegraphics[width=\linewidth]{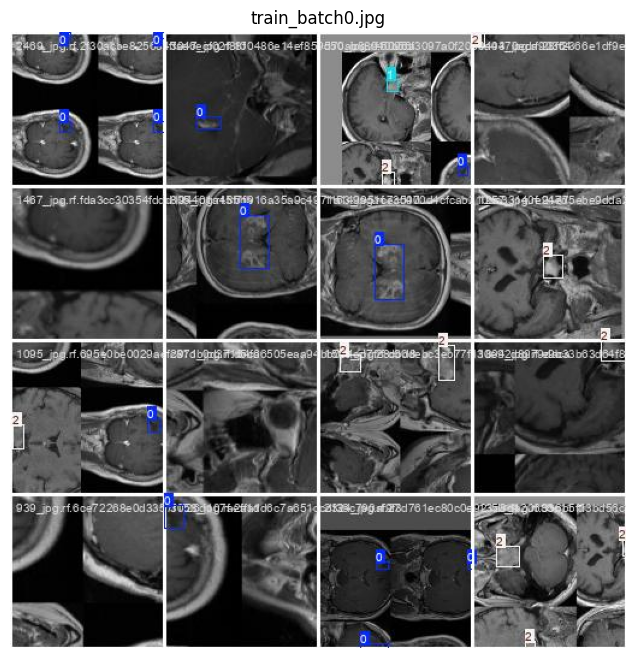}
  \end{subfigure}%
  \begin{subfigure}{0.24\textwidth}
    \includegraphics[width=\linewidth]{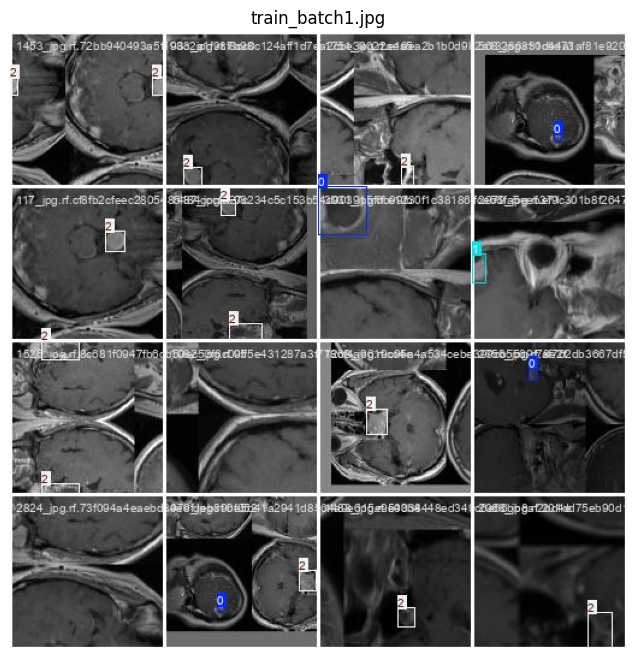}
  \end{subfigure}%
  \begin{subfigure}{0.24\textwidth}
    \includegraphics[width=\linewidth]{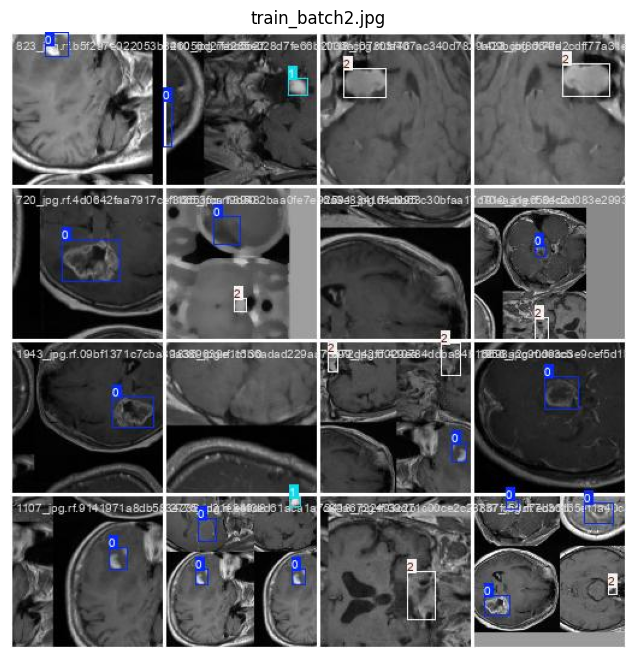}
  \end{subfigure}%
  \begin{subfigure}{0.24\textwidth}
    \includegraphics[width=\linewidth]{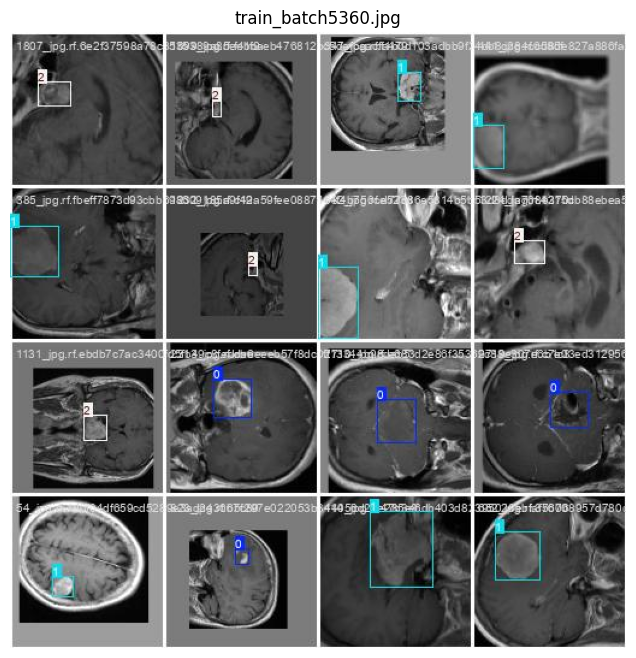}
  \end{subfigure}

  \begin{subfigure}{0.24\textwidth}
    \includegraphics[width=\linewidth]{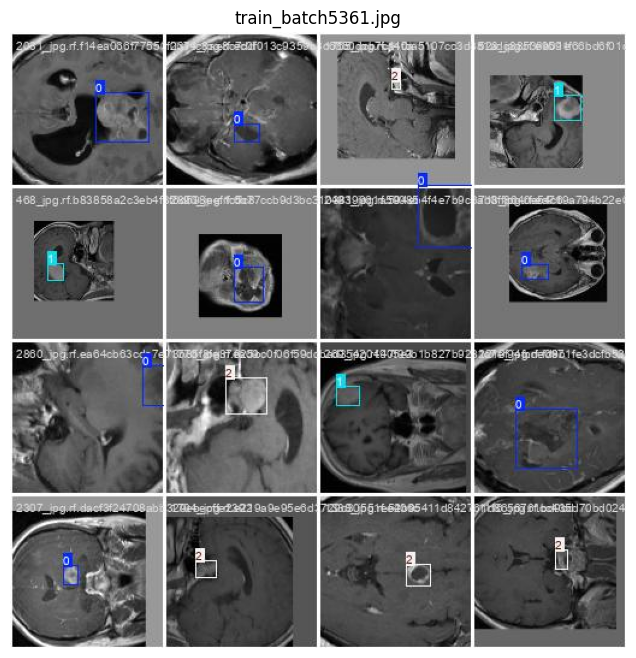}
  \end{subfigure}%
  \begin{subfigure}{0.24\textwidth}
    \includegraphics[width=\linewidth]{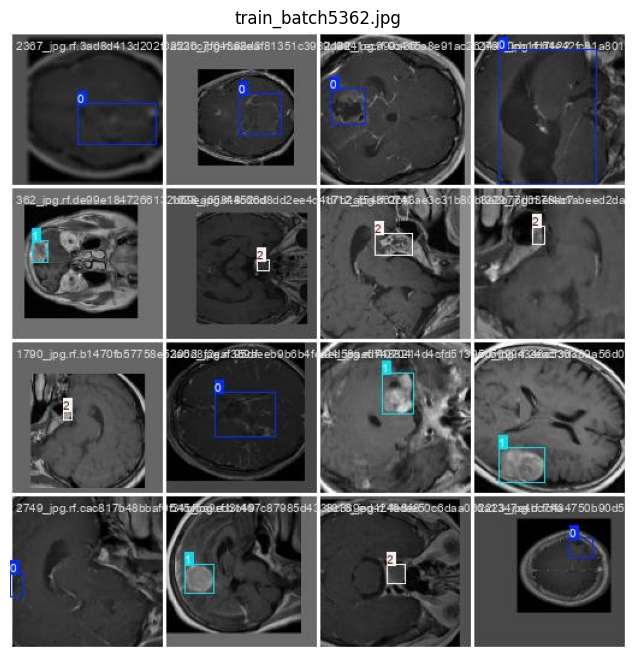}
  \end{subfigure}%
  \begin{subfigure}{0.24\textwidth}
    \includegraphics[width=\linewidth]{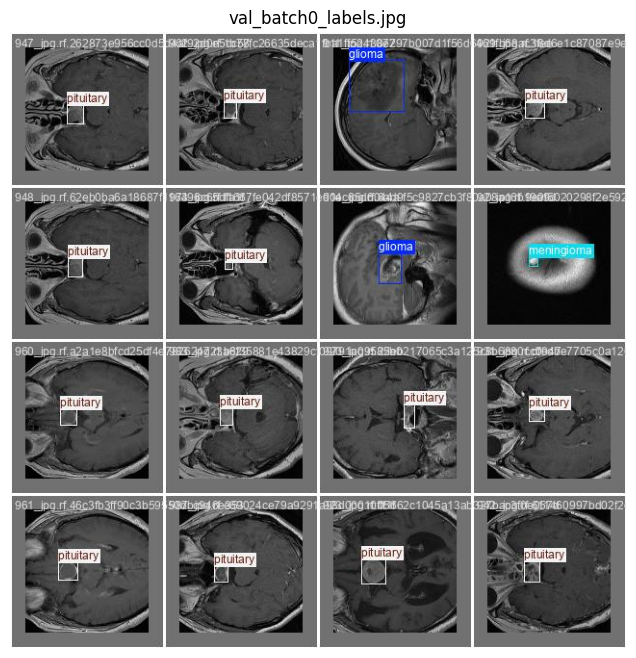}
  \end{subfigure}%
  \begin{subfigure}{0.24\textwidth}
    \includegraphics[width=\linewidth]{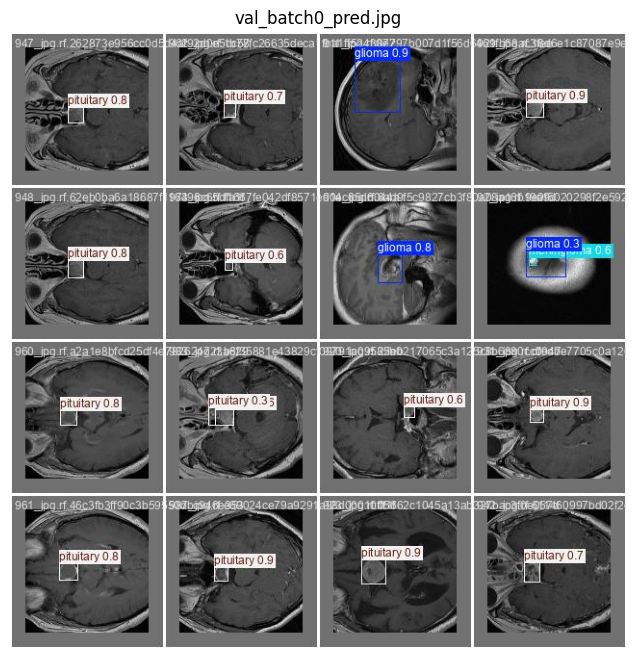}
  \end{subfigure}

  \begin{subfigure}{0.24\textwidth}
    \includegraphics[width=\linewidth]{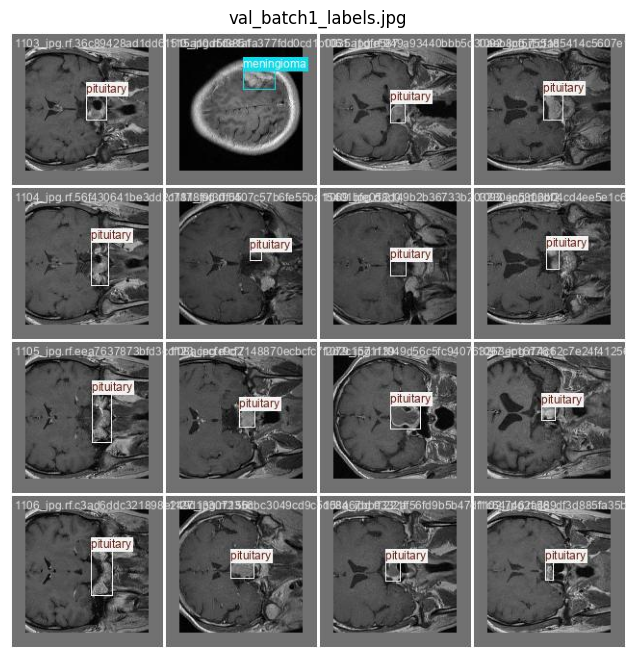}
  \end{subfigure}%
  \begin{subfigure}{0.24\textwidth}
    \includegraphics[width=\linewidth]{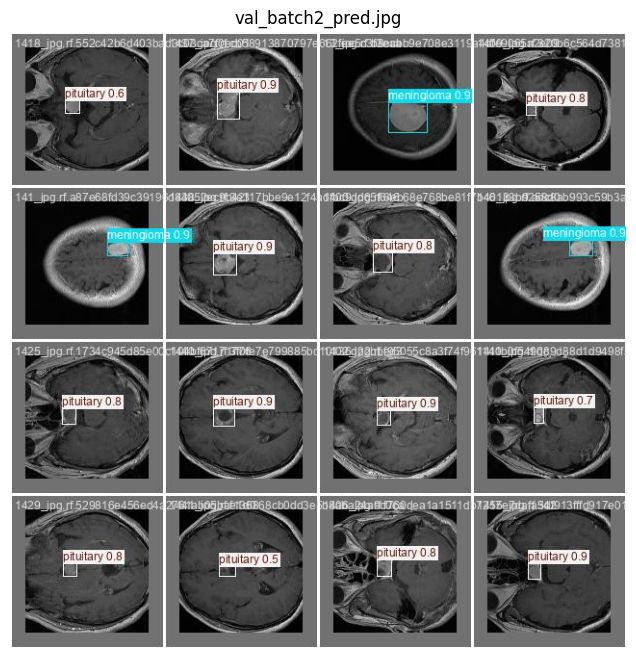}
  \end{subfigure}%
  \begin{subfigure}{0.24\textwidth}
    \includegraphics[width=\linewidth]{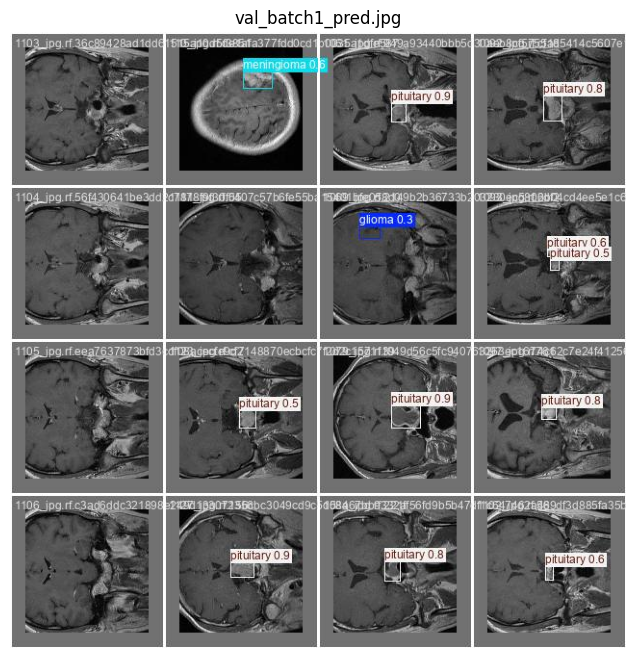}
  \end{subfigure}%
  \begin{subfigure}{0.24\textwidth}
    \includegraphics[width=\linewidth]{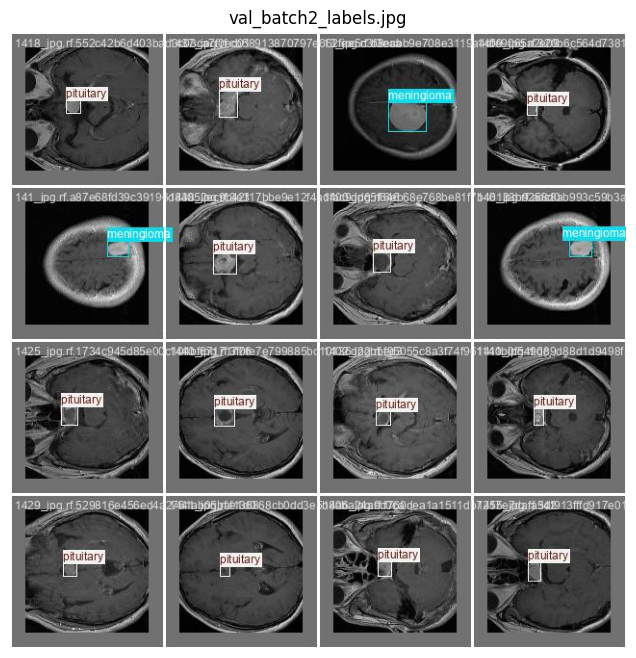}
  \end{subfigure}
    \caption{Simulation results on the model's predictions across various images.}
\end{figure*}
\subsection{Environment settings}
The simulation environment for our federated learning system has been rigorously developed to replicate real-world conditions encountered in medical imaging for tumor detection. We use a simulated network of $N$ clients, each representing a different medical facility and its unique dataset of MRI images with various resolutions and patient profiles. These datasets have been generated synthetically, although they follow realistic brain tumor imaging distributions and characteristics. The server and all clients are simulated on a high-performance computer cluster outfitted with GPUs to parallelize processing and speed up the training procedure. To replicate real-world resource constraints, each client runs its local instance of the training algorithm on a dedicated GPU. The global server, that manages the federated learning process, implements the Federated Averaging method to aggregate model updates. Network latency and capacity are artificially introduced into the simulation to evaluate the model's robustness under typical internet environments. To ensure consistency, software dependencies are standardized across all nodes, and model training and simulation are performed using Python 3.8, TensorFlow 2.x, and PyTorch 1.8. The entire simulation is run under controlled conditions, ensuring reproducibility and enabling the study of the federated learning model's performance over a variety of challenging datasets.

\subsection{Score Matrices}
We simulated four score matrices, such as the f1 score, the p-score, the pr-score, and the r-score in Fig. 2.

\textbf{F1 Score}
The F1 Score is a harmonic mean of precision and recall, balancing the two metrics. It is particularly useful when the costs of false positives and false negatives are similar. The F1 score is calculated as follows:
\[
\text{F1 Score} = 2 \times \frac{\text{Precision} \times \text{Recall}}{\text{Precision} + \text{Recall}}
\]
where \textit{Precision} is the ratio of true positives to the total predicted positives, and \textit{Recall} is the ratio of true positives to the total actual positives.

\textbf{Precision (P-Score)}
Precision, also known as the P-Score, quantifies the accuracy of positive predictions. This metric is crucial where false positives carry a high cost. Precision is defined by the equation:
\[
\text{Precision} = \frac{\text{True Positives}}{\text{True Positives} + \text{False Positives}}
\]
It indicates the correctness of positive identifications by the model.

\textbf{Precision-Recall Score (PR-Score)}
The Precision-Recall Score typically refers to the Precision-Recall curve, which shows the trade-off between precision and recall for different threshold settings of a classifier. The PR curve is essential for understanding a model's performance across various sensitivity levels. While a single metric "PR-Score" is not standard, the concept is crucial for optimizing the threshold of classification models.

\begin{figure*}[ht!]
  \centering
  \begin{subfigure}{0.49\textwidth}
    \includegraphics[width=\linewidth]{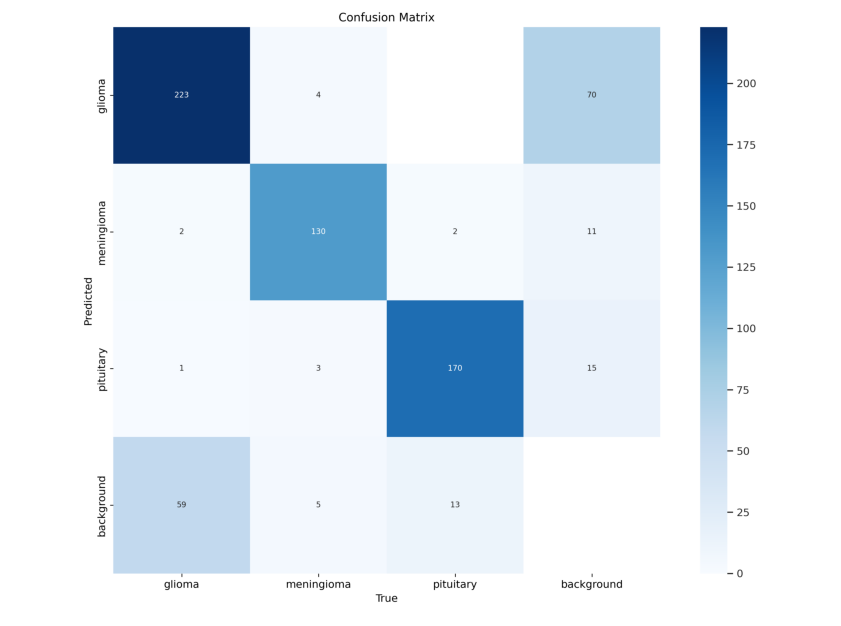}
    \caption{ML Results}
  \end{subfigure}%
  \begin{subfigure}{0.49\textwidth}
    \includegraphics[width=\linewidth]{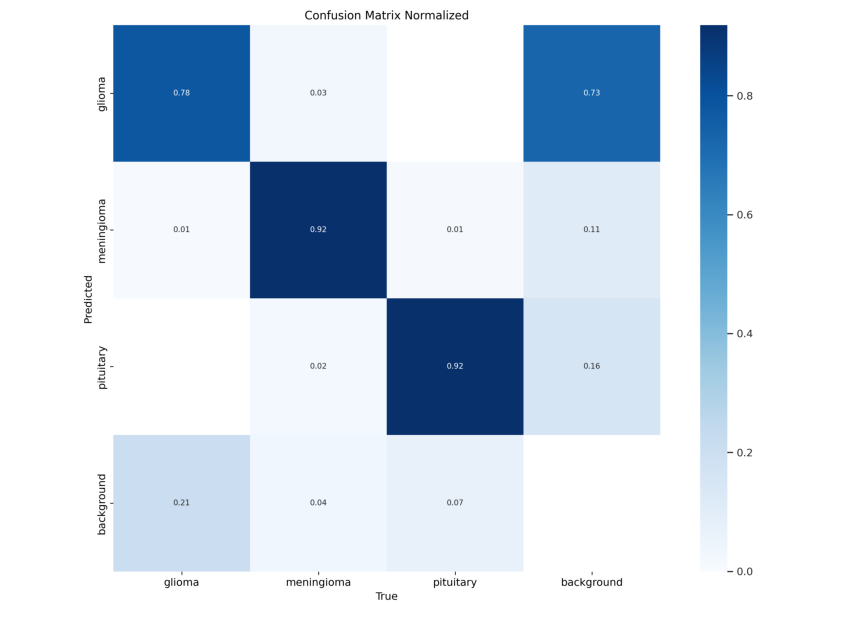}
    \caption{FL Results}
  \end{subfigure}%
  \caption{Comparison between ML (a) and FL (b) results between 2 confusion matrix where the more precise diagonal indicates better model prediction.}
\end{figure*}
\begin{figure*}[ht!]
  \centering
  \begin{subfigure}{0.49\textwidth}
    \includegraphics[width=\linewidth]{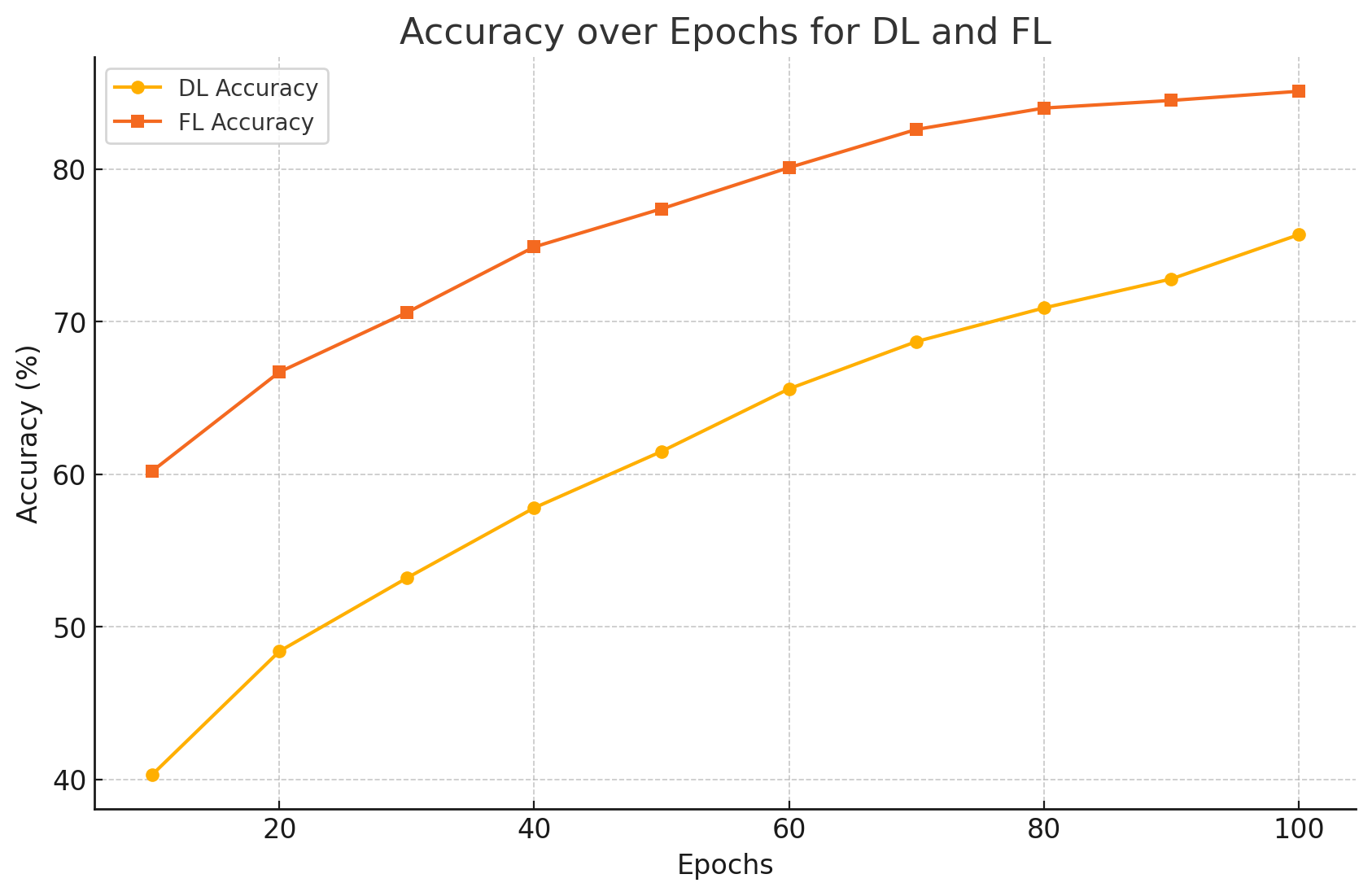}
    \caption{Accuracy }
  \end{subfigure}%
  \begin{subfigure}{0.49\textwidth}
    \includegraphics[width=\linewidth]{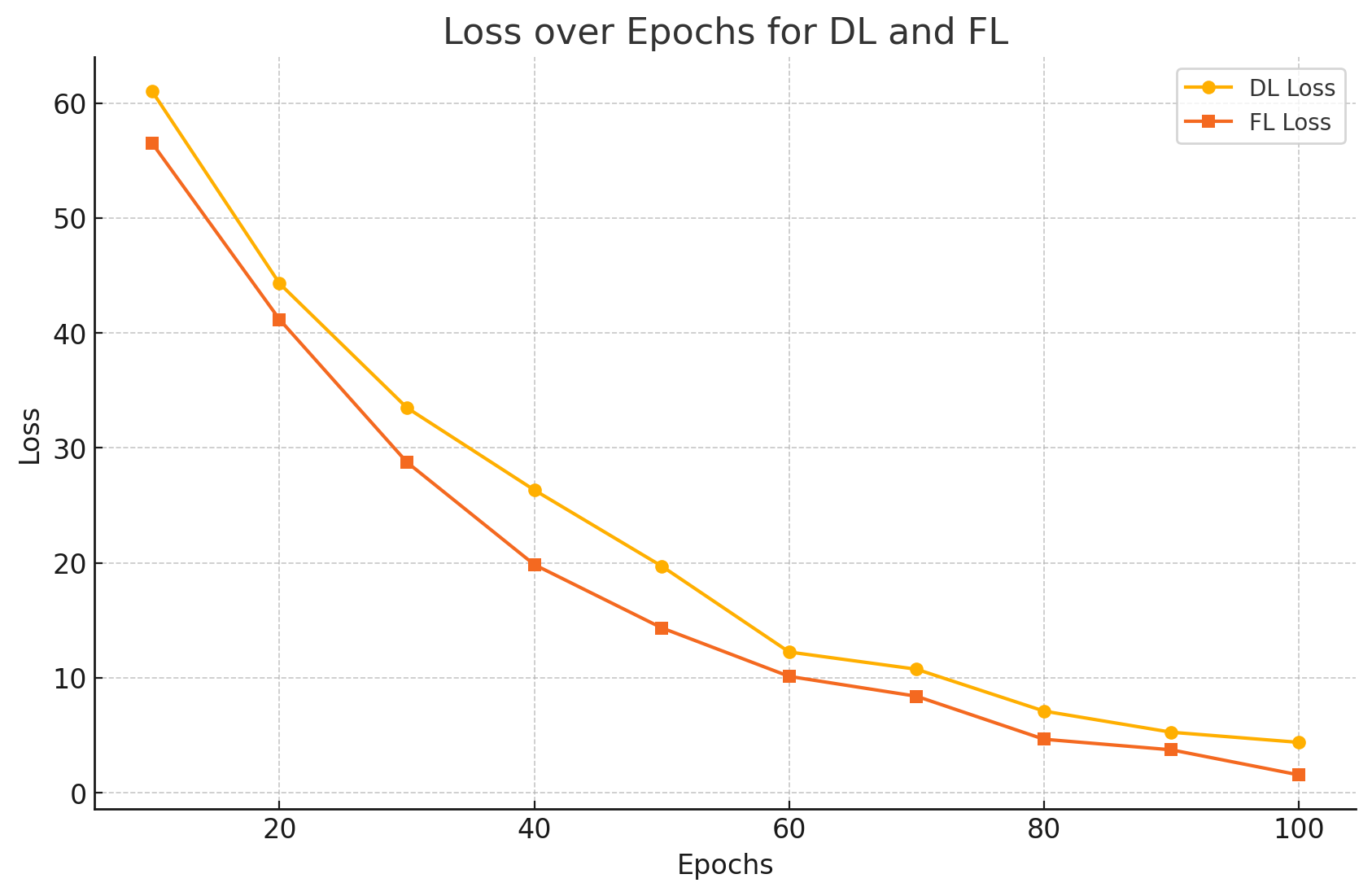}
    \caption{Loss Value}
  \end{subfigure}%
  \caption{Comparison between ML and FL results in terms of accuracy (a) and loss (b) value.}
\end{figure*}
\textbf{Recall (R-Score)}
Recall, or R-Score, evaluates the model's ability to detect all relevant instances. This metric is essential in cases where missing a positive instance is important. Recall is calculated using the following formula:
\[
\text{Recall} = \frac{\text{True Positives}}{\text{True Positives} + \text{False Negatives}}
\]
It reflects the model's sensitivity to identifying positive cases, which ensures the identification process is complete. Glioma had the lowest overall score, followed by all classes and the pituitary. Finally, meningioma is listed top. The results demonstrate that the classification model performs differently across tumor types, with meningioma being the most accurately detected, followed by pituitary tumors and gliomas. This variety suggests that glioma may present features separating issues that are less prominent in other tumor types. Further research into the precise characteristics of gliomas that result in lower scores could aid in the refinement of the model, increasing its accuracy and dependability across all classes. This could consist of increasing the glioma training data, applying more sophisticated feature extraction approaches, or exploring more complex model architectures designed to capture the distinct properties of each tumor type.

\subsection{Brain Tumor Detection Images}
 Figure 3 shows the model's predictions across a variety of images, each presenting either a single kind or a combination of different brain tumor types. The results also provide the model's confidence ratings for each prediction. As a result, this emphasizes the model's ability to not only detect tumors but also distinguish between different tumor types with a high degree of accuracy. This level of prediction confidence is essential for clinical applications that require accurate and reliable diagnostic information. Further improvement of the model could enhance its diagnostic accuracy and confidence levels.

\subsection{Comparison Between ML and FL Approach}
Finally, we compare our proposed FL method with the existing ML method in Fig. 4 and 5. In Fig. 4, we show two confusion matrices. The first confusion matrix in Fig. 4(a) is ML result and the second confusion matrix in Fig. 4(b) is FL result. From the confusion matrix, the FL result is more clear diagonally compared to the ML result. The result in Fig. 5 also shows the same result where in Fig. 5(a), FL has higher accuracy than ML. Similarly, Fig 5(b) shows that the loss value of FL is lower than ML which indicates better performance in FL. 
  
\section{Conclusion}
In conclusion, combining new approaches such as machine learning and deep learning into the field of brain tumor detection provides considerable advancements in medical diagnosis. These technologies, especially the use of YOLOv11 models and federated methods of learning, have the potential to increase the accuracy, speed, and efficiency of brain tumor identification. Furthermore, federated learning deals with essential data privacy and security issues, enabling the use of massive, decentralized datasets while preserving patient confidentiality. The continual development and improvement of these methods will certainly create opportunities for brain tumor detection, diagnosis, and therapy, leading to more precise and personalized medical care.


\ifCLASSOPTIONcaptionsoff
  \newpage
\fi

\bibliography{main}
\bibliographystyle{IEEEtran}

\end{document}